\documentclass{article}

\usepackage[final]{corl_2020} 

\usepackage[utf8]{inputenc} 
\usepackage[T1]{fontenc}    
\usepackage{hyperref}       
\usepackage{url}            
\usepackage{booktabs}       
\usepackage{amsfonts,amsmath,mathtools}       
\usepackage{nicefrac}       
\usepackage{microtype}      
\usepackage{subcaption}
\usepackage[font=small]{caption}
\usepackage{float,wrapfig}
\usepackage{multirow}
\usepackage{todonotes}
\usepackage[subtle]{savetrees}

\newcommand{\mypara}{\vspace{-3mm}\paragraph}
\title{Learning a Decentralized Multi-arm Motion Planner}

\author{
  Huy Ha, \quad Jingxi Xu, \quad Shuran Song \\ 
  Columbia University, New York, NY, United States \\
  \url{https://multiarm.cs.columbia.edu/}
}

\begin{document}

\maketitle
\vspace{-3mm}
\begin{abstract}
    We present a closed-loop multi-arm motion planner that is scalable and flexible with team size.
    Traditional multi-arm robot systems have relied on centralized motion planners, whose runtimes often scale exponentially with team size, and thus, fail to handle dynamic environments with open-loop control.
    In this paper, we tackle this problem with multi-agent reinforcement learning, where a decentralized policy is trained to control one robot arm in the multi-arm system to reach its target end-effector pose given observations of its workspace state and target end-effector pose.
    The policy is trained using Soft Actor-Critic with expert demonstrations from a sampling-based motion planning algorithm (i.e., BiRRT).
    By leveraging classical planning algorithms, we can improve the learning efficiency of the reinforcement learning algorithm while retaining the fast inference time of neural networks.
    The resulting policy scales sub-linearly and can be deployed on multi-arm systems with variable team sizes.
    Thanks to the closed-loop and decentralized formulation, our approach generalizes to 5-10 multi-arm systems and dynamic moving targets (>$90\%$ success rate for a 10-arm system), despite being trained on only 1-4 arm planning tasks with static targets.  Code and data links can be found at \href{https://multiarm.cs.columbia.edu}{https://multiarm.cs.columbia.edu}.
\end{abstract}

\section{Introduction}

Many complex manipulation tasks can be decomposed into smaller sub-tasks and distributed amongst multiple robotic arms working in parallel in a shared workspace.
However, efficiently motion planning for such multi-arm systems remains a challenge due to its high degrees-of-freedom (DoF) and tightly coupled workspaces.

While traditional \textit{centralized} motion planner~\citep{lavalle1998rapidly,kurosu2017simultaneous, berg2005prioritizedmultirobot, mirrazavi2018unified,solovey2013drrt, salzman2016multilinkdrrt} benefit from having access to all the information a motion planner module might need, these approaches fail to scale efficiently with the number of arms in the system (the team size) because their centralized components can become the bottleneck of the system.
This scalability issue has limited multi-arm applications requiring large numbers of robotic arms operating in a tight workspace or in dynamic environments with moving targets.

\begin{figure}[t]
      \centering
      \includegraphics[width=\textwidth]{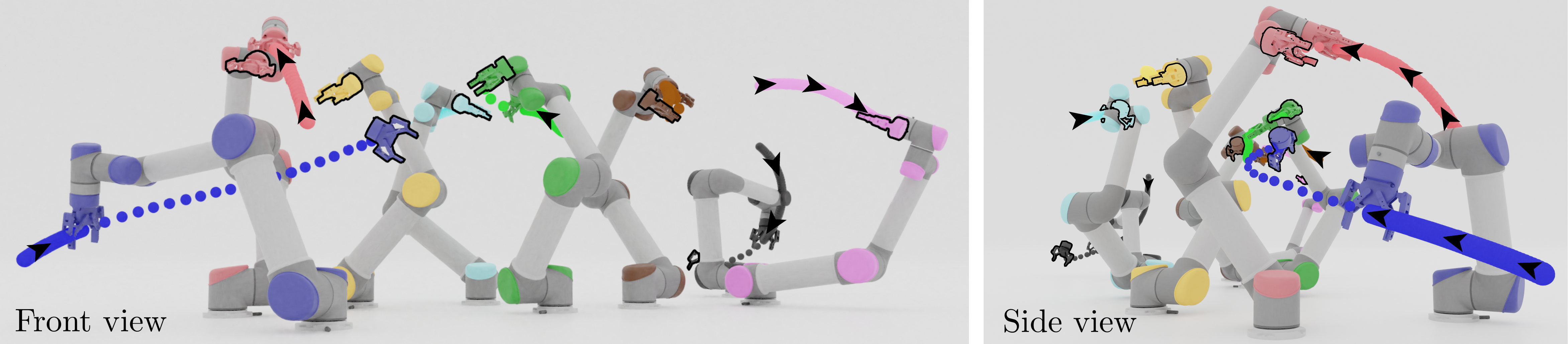}
      \caption{\textbf{Multi-arm Motion Planning.} Our motion planner generalizes to tightly coupled multi-arm systems with any number of arms, while being trained on only 1-4 arms. Here is the motion plan generated by our algorithm (past trajectory is solid line, future trajectory is dotted line) for an 8-arm system to reach their target end-effector poses (outlined black). }
      \label{fig:teaser}
      \vspace{-5mm}
\end{figure}

Less explored alternatives are \textit{decentralized} motion planners, which treat the multi-arm system as a multi-agent system.
Here, each arm is controlled by an agent that receives as input a partial observation of the system's state and computes a motion plan for only \textit{itself}.
Naturally, decentralized motion planners scale efficiently, but designing such a controller for a task as complex as generic multi-arm motion planning remains a challenge.
Through observations of other arms' states alone, decentralized motion planners must efficiently coordinate to avoid collisions and cooperate to collectively reach their target end-effector poses, all the while having control over only its arm (Fig.~\ref{fig:teaser}).
An ideal candidate for a multi-arm motion planner should have the following characteristics:
\begin{itemize}
      \item \textbf{Scalability}:
            The runtime should scale efficiently with the number of arms in the system.

      \item \textbf{Cooperative}:
            Each arm must both reach its target and ensure that it does not prevent other arms from reaching their targets by colliding with other arms or acting as an obstacle.

      \item \textbf{Closed-Loop}:
            The motion planner must react quickly to the unknown dynamics of task-relevant objects in the environment and regenerate its motion.
            Therefore, to avoid redundant computation, a closed-loop motion planner computes just the next waypoint to reach which brings the arm closer to its target rather than the entire trajectory of waypoints which brings the arm to its target.

      \item \textbf{Flexibility}:
            The planner should work for any number of arms and any system layout without the need for a lengthy recomputation or retraining.
            This characteristic is crucial for multi-arm applications which are structurally dynamic, where arms are frequently added to, removed from, or moved around the workspace.
\end{itemize}

We use multi-agent reinforcement learning (MARL) to learn a decentralized motion planning policy, which is incentivized to be cooperative with a team reward when all arms have reached their targets.
The resulting sparse reward problem prevents the random exploration of typical reinforcement learning algorithms to converge to a successful policy.
Additionally, behavior cloned policies from a sampling-based motion planning expert (e.g.,  Bidirectional Rapidly-exploring Random Trees, BiRRT) also perform poorly because they do not observe enough negative examples to learn accurate collision boundaries, which are required for policies to be robust in near-collision states. 

To address these challenges, we propose to supply expert demonstrations (i.e., centralized BiRRT) on failed tasks only, along with the policy's own failed attempt, which has the following advantages.
First, by directly contrasting failed and successful trajectories for the same task, the policy can escape local minima while still experiencing enough unsuccessful cases to be robust.
Second, as the policy improves during training, it requires less expert demonstrations and starts to collect successful trajectories outside of expert demonstrations through self-exploration, which allows it to improve beyond the expert.
Finally, by approximating an expensive expert (i.e., centralized BiRRT) with a fast neural network, the final system retains fast inference speeds at test time, which allows it to quickly react to the unknown dynamics of targets and replan.
Our motion planning policy achieves high success rates in tightly coupled multi-arm static and dynamic motion planning tasks, something which neither sampling-based nor learning-based approaches can achieve alone.

Our main contribution is an MARL approach for training a cooperative, closed-loop, decentralized multi-arm motion planner that scales sub-linearly with the number of arms in the system, and can plan at a rate of 920Hz on a single CPU thread.
Our policy uses a Long Short-Term Memory (LSTM)~\citep{hochreiter1997long} module to encode a variable number of arms, which allows it to be flexible with respect to the team size.
Despite being trained on only 1-4 arm \textit{static} motion planning tasks, our approach generalizes to 5-10 arm static and \textit{dynamic} motion planning tasks in test time.

\section{Related Work}
\vspace{2mm}
\mypara{Sampling-based Motion Planning.}
Sampling-based motion planning (SMP) algorithms such as Rapidly-exploring Random Trees (RRT)~\citep{lavalle1998rapidly, lavalle2006planning}, optimal Rapidly-exploring Random Trees (RRT*)~\citep{karaman2011sampling}, Potentially
Guided RRT* (P-RRT*)~\citep{qureshi2016potential} and their bi-directional variants~\citep{tahir2018potentially, qureshi2015intelligent} have long been the go-to solutions for generic motion planning, due to their efficiency in practice and completeness guarantee in theory. 
However, they struggle to scale to high-dimensional settings that are common in many real-world applications because their runtime is exponential in the systems's DoF, which is the sum of individual robots' DoF in multi-robot systems.
A faster alternative to RRT in runtime is Probabilistic Road Maps (PRM)~\citep{kavraki1996prm}, however, generating sufficiently dense roadmaps for large multi-robot systems is costly and not flexible.

\vspace{1mm}
\mypara{Learning-based Motion Planning.}
Many recent works have proposed learning-based approaches, which trade-off a lengthy offline training stage for faster online execution. 
Some of these works have improved the speed of SMPs by biasing the sampling distribution of new nodes~\citep{ichter2017samplingdistributionrrt,qureshi2018deeply, jetchev2013fast}, directly predicting the next node~\citep{qureshi2019motion}, utilizing past experience with an extra lookup table module~\citep{berenson2012robot}, or reducing the search space through discretization~\citep{solovey2013drrt,salzman2016multilinkdrrt,shome2019drrtstar}.
Others have explored the application of reinforcement learning to motion planing~\cite{zhang2018auto,pfeiffer2017perception,bency2018towards,faust2018prm, tamar2016value}. For example, \citet{jurgenson2019rlmotionplanning} combined Deep Deterministic Policy Gradient~\citep{lillicrap2015continuous} with expert demonstrations from RRT* for planar motion planning of a 4-DoF robot arm. However, as these approaches do not explicitly account for multiple robots, they can only operate in a centralized fashion, and therefore is not scalable.

\vspace{1mm}
\mypara{Centralized Multi-robot Motion Planning.} 
The multi-arm motion planning problem has been studied in applications such as pick and place~\cite{shome2019anytime} and table-top rearrangement~\cite{shome2018tabletop}.
\citet{mirrazavi2018unified} proposed a fast closed-loop multi-arm controller capable of replanning at 500Hz for dual-arm systems, but relies on a centralized Inverse Kinematics (IK) solver, whose constraints must be learned from scratch if arms are added to, removed from, or moved around the workspace.
Berg et al.~\citep{berg2005prioritizedmultirobot} explored a prioritized planning approach for multi-robot systems, but used \citep{berg2005roadmapdynamic} for motion planning which is open-loop and requires a lengthy computation of roadmaps for each specific workspace.
Most importantly, these approaches suffer from poor scalability because they have centralized components that may act as bottlenecks in the system. 

\vspace{1mm}
\mypara{Decentralized Multi-robot Motion Planning.}
To achieve true scalability, motion planners should be decentralized, where each robot plans to reach its target configuration.
Non-Holonomic Optimal Reciprocal Collision Avoidance (NH-ORCA) \cite{alonso2013optimal} is a popular approach for distributed multi-robot planning and collision avoidance, but is sensitive to hyper-parameters.
Other MARL prior works for multi-robot motion planning \cite{everett2018ga3ccadrl,semnani2020multi} have only investigated single-link, low dimensional configuration space planning of ground robots, and do not work for our tightly coupled multi-arm systems with 6-DoF robot arms (see discussion in supplementary material).


\section{Background}
\vspace{2mm}
\mypara{Problem Definition.}
The goal of a multi-arm motion planning task is to find a collision-free trajectory of waypoints for all arms from their initial joint configurations to reach their target end-effector poses.
Let $i$ be an arm with a stationary base pose $p_i^b$ and a joint configuration space $\mathcal{C}_i \subseteq \mathbb{R}^d$, where $\mathcal{F}_i \subseteq \mathcal{C}_i$ is the free space of $i$ -- the set of all configurations for which $i$ is not in collision.
We say $i$ has reached its target end-effector pose $p_i^t$ (from now on referred to as its target pose) if its current end-effector pose $p_i^e$ is within a distance of $\epsilon_{p}$ and a rotation of $\epsilon_{r}$ away from $p_i^t$.
A multi-arm motion planning problem for $\mathcal{N}$ arms is described by the arms' base poses, initial joint configurations, and target poses.
Let $\mathcal{F} = \mathcal{F}_1 \times \mathcal{F}_2 \times \dots \times \mathcal{F}_\mathcal{N}$ be the composite free space of all arms.
A solution to a multi-arm motion planning problem is the continuous function $\sigma : [0,1] \rightarrow \mathcal{F}$, such that $\sigma(0)$ is composite initial joint configuration and $\sigma(1)$ is a configuration where all arms reach their target poses.

\mypara{Multi-agent Reinforcement Learning (MARL).}
The extension of Markov Decision Processes to multi-agent scenarios are Partially Observable Markov Games (POMGs).
A POMG for $\mathcal{N}$ agents has a set of states $\mathcal{S}$ describing the environment and all possible configurations of all agents, a set of action spaces $\{\mathcal{A}_i\}_{i\in \mathcal{N}}$ and a set of observation spaces $\{\mathcal{O}_i\}_{i\in \mathcal{N}}$.
Agent $i$ uses a stochastic policy $\pi_{\theta_i} : \mathcal{O}_i \times \mathcal{A}_i \rightarrow [0,1]$ parameterized by $\theta_i$ to choose an action from its observation.
The joint action of all agents produces the next state according to the transition function $\mathcal{T}: \mathcal{S} \times \mathcal{A}_1 \times \dots \times \mathcal{A}_N \rightarrow \mathcal{S}$.
Agent $i$ then receives the next partial observation correlated with the state $o_i: \mathcal{S} \rightarrow \mathcal{O}_i$ and a scalar reward from its own specific reward function $\mathcal{R}_i: \mathcal{S} \times \mathcal{A}_1 \times \dots \times \mathcal{A}_N \times \mathcal{S} \rightarrow \mathbb{R}$.
All agents' goals are to maximize their own total expected returns $R_i=\sum_{t=0}^T \gamma^t r^t_i$, where $\gamma$ is the discount factor and $T$ is the time horizon.
\section{Approach: Decentralized MARL with Expert Demonstrations}\label{sec:method}

We propose a method to train a cooperative, closed-loop, decentralized multi-arm motion planner which can compute motion plans for multi-arm systems of arbitrary team sizes to reach both static and dynamic target poses.
We use Independent Learning \cite{tan1993multi} with Soft Actor Critic \cite{haarnoja2018sac}, where each arm in the multi-arm system is controlled by an agent in the homogeneous, cooperative multi-agent system using a decentralized policy with shared weights.
Our low-level motion planning policy can be combined with any high-level task planner to achieve task-level heterogeneity for any multi-arm team size where arms have the same hardware (Section \ref{section:demo}).

\vspace{-1mm}
\subsection{Generating Multi-arm Motion Planning Tasks}\label{sec:taskgeneration}
\vspace{-1mm}
Each motion planning task we generate can be used in two modes: (1) static mode, where target poses are stationary, and (2) dynamic mode, where target poses move at a constant speed in the range of $1-15$ cm/s.
Our training and testing tasks are generated using the same procedure as follows.

First, for a task for $k$ arms, sample $k$ random planar base poses.
Then, after setting the $k$ arms to the $k$ base poses, generate 3 collision free composite joint configurations $[q_1,q_2,q_3]$ by solving IK for 3 random target end effector positions, with a bias towards the home configuration.
The resulting joint configurations are not uniform in the end effector orientations, but nonetheless lead to tight workspaces.
In static mode, use $q_1$ as the initial joint configuration, $q_2$ to compute the target end-effector poses using forward kinematics (FK), and don't use $q_3$.
In dynamic mode, use $q_1$ as the initial joint configuration, and interpolate between $q_2$ and $q_3$ over the course of the episode, computing the FK solutions of the current interpolated joint configurations to get the current dynamic target poses.
This approach generates tasks which can be used for both static and dynamic motion planning while ensuring the existence of a solution for all tasks.

The training dataset consists of 1,000,000 tasks in static mode, split evenly among 1-4 arms, and the testing dataset consists of 30,000 tasks in both static and dynamic modes, split evenly among 1-10 arms.

\vspace{-1mm}
\subsection{Multi-agent Reinforcement Learning Formulation.}\label{sec:marlsetup}
\vspace{-1mm}

We formulate the multi-arm motion planning task as a homogeneous, cooperative multi-agent reinforcement learning problem, where each arm is controlled by an instance of \textit{ the same policy} with weights shared across all arms.
The agents were trained using Independent Learning (IL) \cite{tan1993multi} with Soft Actor Critic \cite{haarnoja2018sac}, where agents share parameters and experience.

To achieve cooperation, we use a team reward, which rewards all arms $+1$ only when all arms have reached their corresponding target end-effector poses.
Using this team reward and allowing each arm to observe other arms' states in its local workspace (Sec.~\ref{section:vlo}), arms are motivated and given enough information to not block other arms from other arms' targets.
Additionally, arms receive a reward of $+0.01$ when it individually reaches its target and a penalty of $-0.05$ when it collides.
The policy is trained with a curriculum which decreases $\epsilon_p$ and $\epsilon_r$ over time (see supplementary materials for details).

In each episode, a task from the training dataset is sampled.
The episode terminates when one of the arms collide, when all arms reach their targets, or when the episode has reached 500 time steps, then the experiences of all arms are collected into one replay buffer used to optimize the policy.
The task is marked as successful if all arms reach their targets and failed otherwise.

\begin{figure}[t]
    \centering
    \includegraphics[width=\textwidth]{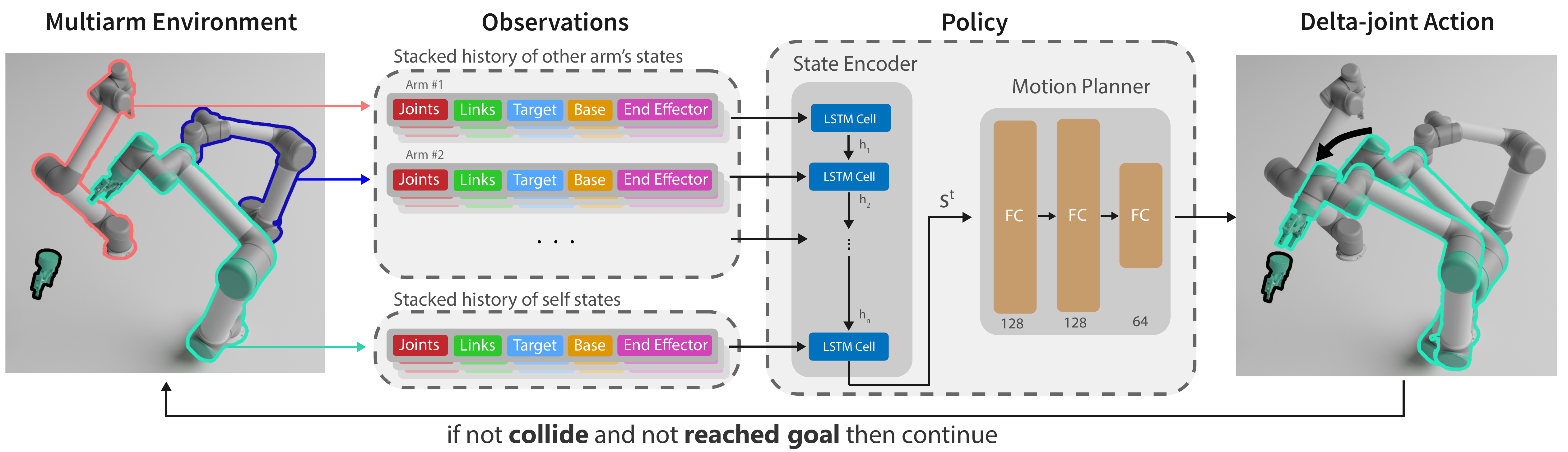}
    \caption{\textbf{Variable Team Size with LSTM State Encoder.}
        From the multi-arm environment, a local workspace embedding $s^t$ is extracted from the variable-length sequence of stacked histories of an arm's self (outlined mint green) and other arms' (outlined pink and blue) states.
        The workspace embedding $s^t$ is used by a Motion Planner multi-layer perceptron, which outputs a delta-joint action (black arrow) that brings the arm closer to its target end-effector pose (outlined black).}
    \vspace{-4mm}
    \label{fig:method}
\end{figure}

\vspace{-1mm}
\subsection{Team Size Flexibility with LSTM State Encoder}\label{section:vlo} 
\vspace{-1mm}

In multi-arm systems, the dimensionality of the state space increases linearly with the team size.
However, training one policy for each team size is costly and not scalable.
To address this issue, we propose to use a Long Short Term Memory (LSTM) \citep{hochreiter1997long} as a state encoder to extract a fixed-sized embedding of an arms' local workspace from the variable length sequence of arm states,
which allows our policy to achieve sub-linear runtime and team-size flexibility.
Unlike masked observations as used in \cite{baker2019emergent}, our LSTM state encoder does not have a maximum limit on how many agent states can be encoded.

Each arm's state consists of its base pose, joint configuration, and target end-effector pose.
To ease the learning process, we additionally include into each arms' state its end-effector pose and link positions, and stack 1 frame of past histories of all state components except for its base pose (Fig. \ref{fig:method}).

At each time step, each arm observes only arms whose base positions are within a 85cm radius from its own.
We process the variable length sequence of arm states (i.e. stacked histories) using an LSTM state encoder to extract a 256 dimensional embedding $s^t$.
The sequence is sorted in decreasing arms' base distance to the current arm, so that an arm's self-observation is always the last element in the sequence.

After $s^t$ is extracted, it is passed through $3$ fully connected layers with hyperbolic tangent activations to output a $12$ dimensional vector.
This vector parameterizes a $6$ dimensional Gaussian action distribution, from which the arms' delta-joint actions can be sampled.
The $Q$ function has the same architecture as the policy, except for the last layer, which has an output dimension of $1$ and no activation function.

\vspace{-2mm}
\subsection{Expert Demonstrations on Failed Motion Planning Tasks}\label{section:expert}
\vspace{-1mm}
\begin{figure}[t]
    \centering
    \includegraphics[width=\textwidth]{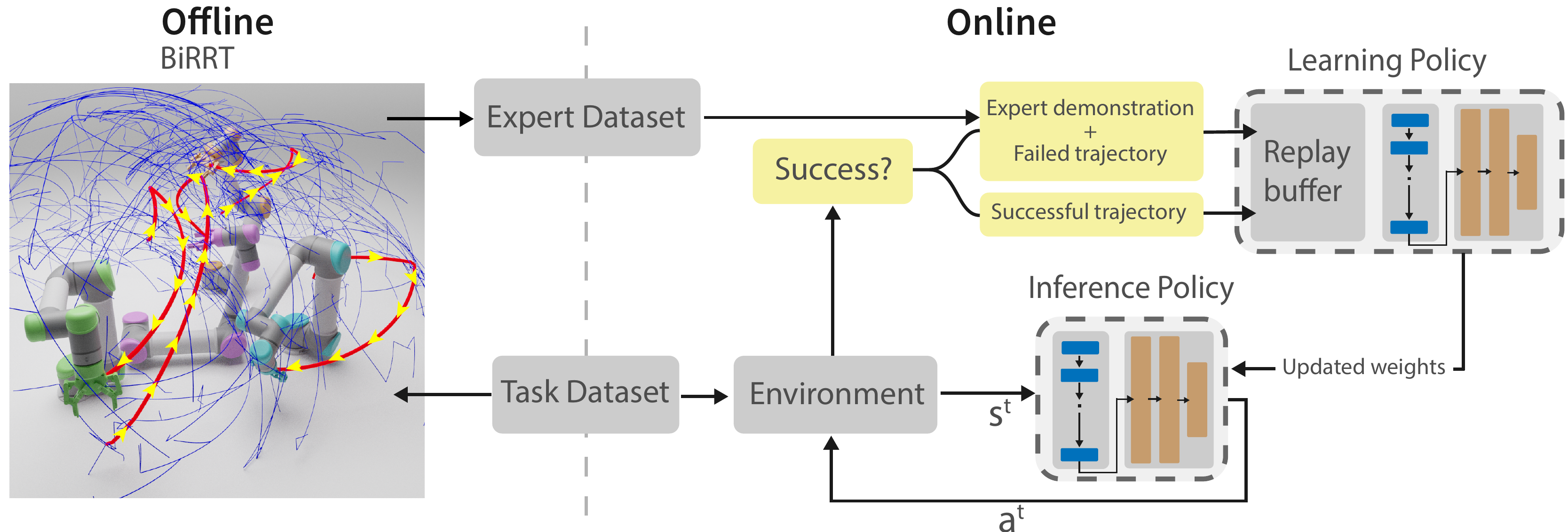}
    \caption{\textbf{Learning with Expert Demonstrations.} If the RL policy fails at a task, the expert demonstration, computed from BiRRT offline, is supplied, along with the policy's failed trajectory, to guide the policy's exploration to higher-value actions. Our experiments demonstrate that this hybrid training scheme is critical for learning a successful policy in complex tasks with sparse rewards. }
    \vspace{-4mm}
    \label{fig:expertdemonstrations}
\end{figure}

While the proposed team reward incentivizes team coordination, it also leads to sparse rewards, where random exploration can not achieve enough positive feedback.
To guide initial exploration and help the policy escape local minima, we supply successful centralized BiRRT trajectories to the policy on failed tasks, which addresses the sparse reward problem while still allowing the policy to learn from failed experiences.
Additionally, this design is robust against expert's optimality, and converges to successful policy as long as expert trajectories are successful.

Before training, a centralized BiRRT expert computes a dense trajectory
for tasks in the training task dataset.
During training, if the policy fails at a task, the precomputed BiRRT waypoints for that task are converted to individual arms' delta-joint actions and executed.
Successful expert experiences are then added to the replay buffer, along with the policy's own failed experience (Fig. \ref{fig:expertdemonstrations}).

The number of waypoints in BiRRT's dense trajectories may exceed the time step limit, so computing expert delta-joint actions by subtracting consecutive waypoints does not allow the arms to reach their targets in time.
%
Instead, we treat the trajectory as a curve composed of line segments in joint space, and decimate it with an angle tolerance factor of 0.01 radians to get a simplified trajectory.
Delta-joint actions are then found by subtracting the arms' current configuration from the next waypoint in the simplified trajectory, with an upper limit on the action magnitude of 0.5 on all joint dimensions.
This generates a sequence of delta-joint actions which approximates BiRRT's original trajectory with enough details to avoid collisions while still reaching the target in time.

For failed tasks, contrasting the policy's low-reward failed attempts with the expert's high-reward successful experience helps the policy during the exploration phase and guides it out of local minima.
For successful tasks, expert trajectories are not supplied, giving the policy room to improve beyond the expert through self-exploration.
Therefore, as long as the expert is successful, the actual optimality of the expert does not affect the final optimality of the policy, because the agent can continue to refine its optimality through successful trajectories outside of expert demonstrations collected by self exploration.

Since the replay buffer contains experience from both the current policy and the expert, we choose Soft Actor Critic \cite{haarnoja2018sac} for being an off-policy RL algorithm which learns an entropy maximizing stochastic policy for continuous action spaces.

\section{Evaluation}
\vspace{2mm}
\mypara{Success Metric.}
A task is successful if there is a time step in the episode in which all arms have reached their target end-effector poses within a distance of 2cm and a rotation of 0.1 radians.
Alternatively, if any arm in the system collides, or 500 time steps have passed, then the task is unsuccessful.

\begin{figure}[t]
  \includegraphics[width=1.0\textwidth]{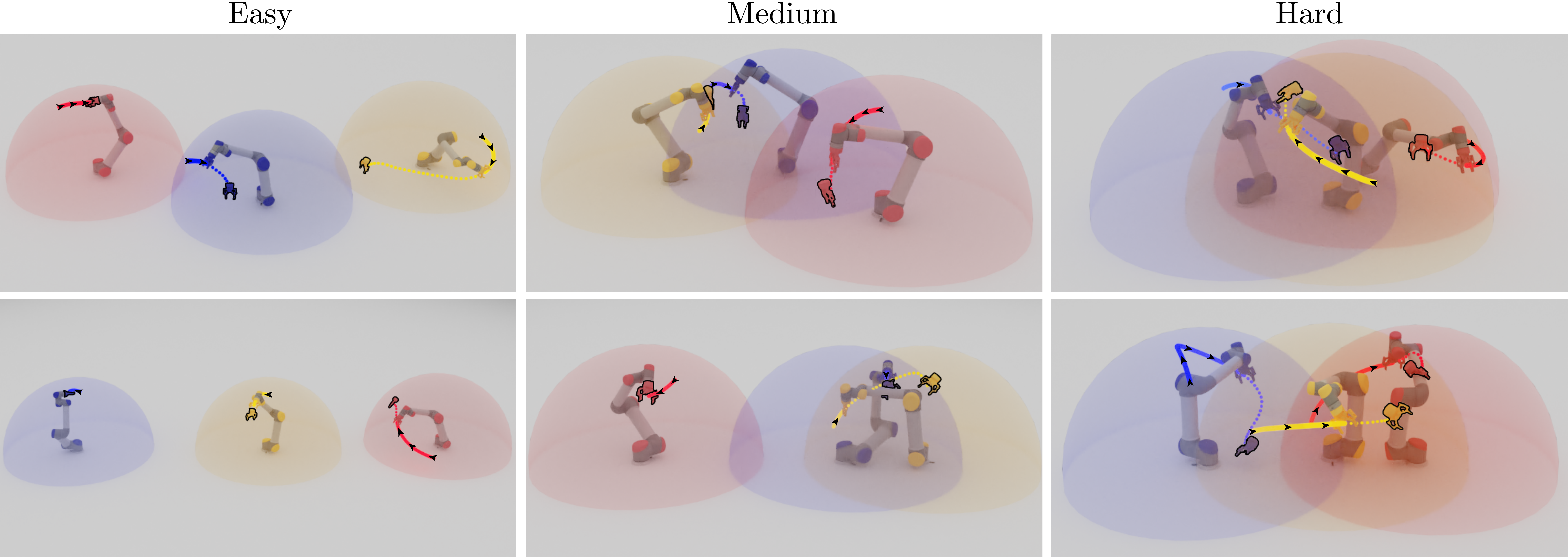}
  \caption{
    \textbf{Motion Planner Results with Different Difficulty Levels.}
    Each task's difficulty is measured as the maximum arm's workspace \% volumetric intersection with other arms' workspaces (semi-transparent hemispheres).
    Intuitively, easy tasks can be reduced to multiple single-arm planning tasks, while hard tasks involve significant overlap between the arms' workspaces.
    Our approach can find collision-free trajectories (solid lines for past trajectories, dotted lines for future trajectories) to reach target end-effector poses (outlined black) with high success rates for all three difficulty levels.
  }
  \label{fig:difficulty}
  \vspace{-4mm}
\end{figure}

\vspace{1mm}
\mypara{Task Difficulty.}
Tasks are divided up based on the difficulty, which is a measure of how tightly coupled the workspace is, and thus, how likely a collision is between arms.
A task's difficulty is the maximum \% intersection of any arm's hemispherical workspace volume (visualized in Fig. \ref{fig:difficulty}) with all other arm's hemispherical workspace volumes.
Taking the maximum makes tasks with different team sizes comparable (compared to taking the sum), while still being preferable to taking the mean since a task is unsuccessful if any arm in the system is unsuccessful.

\vspace{1mm}
\mypara{Testing Dataset.}
The testing dataset consists of 30,000 tasks with novel base poses and target end effector poses, split evenly between 1-10 arms, and generated in the same manner as the training dataset (Sec.~\ref{sec:taskgeneration}).
Static tasks are divided based on their difficulties, where easy, medium, and hard tasks have difficulty range $0.0 - 0.35$, $0.35-0.45$, and $0.45-0.50$ respectively.
Dynamic tasks are instead divided based on their speeds, where slow, medium, and fast tasks have difficulty range $1-5$ cm/s, $5-10$ cm/s, and $10-15$ cm/s respectively, with an average difficulty of 0.185.

\vspace{1mm}
\mypara{BiRRT Implementation.}
For computing the expert demonstration dataset and testing,
our implementation of BiRRT uses euclidean joint space distance as the distance function, and is ran for 4000 iterations at a resolution of 0.05 radians.
For static tasks, BiRRT uses the initial $q_1$ and goal $q_2$ configuration from the task (Sec.~\ref{sec:taskgeneration}).
For dynamic tasks, BiRRT uses the multi-arm system's current joint configurations as its initial configuration, and the current interpolation between $q_2$ and $q_3$ as described in Sec.~\ref{sec:taskgeneration} as its goal configuration.
The system is allowed to execute the entire BiRRT trajectory before a new computation starts with the latest start and goal configurations as input.
For static tasks, BiRRT's runtime is measured from when sampling begins to when a collision-free trajectory for all arms is found.

\vspace{1mm}
\mypara{Training.}
On a system with the 16 core Intel i7-7820X CPU and an NVIDIA GTX 1080 GPU, training our policy take 668.2 M timesteps (approx. 14 days or 700,000 episodes), and generating 1,000,000 expert waypoints takes 2 days.
We use PyBullet \cite{coumans2017pybullet} as our training and testing environment.
\newpage
\subsection{Experiment results} \vspace{-1mm}

\begin{wrapfigure}{r}{.45\textwidth}
  \centering
  \vspace{-7mm}
  \includegraphics[trim=55   0   90   10,clip,width=0.45\textwidth]{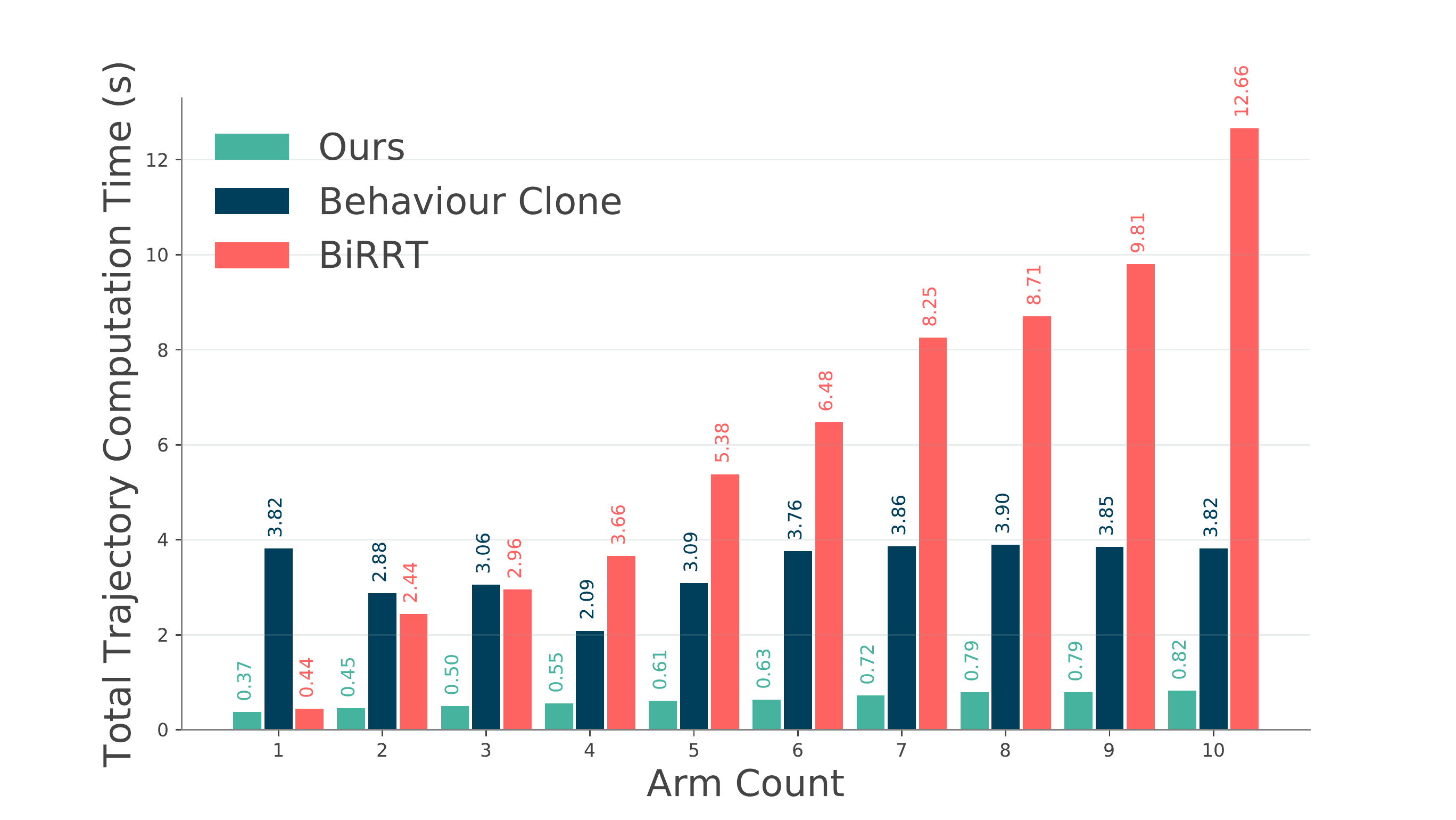}
  \vspace{-6mm}
  \caption{Runtime Speed and Scalability.
  }
  \vspace{-4mm}
  \label{fig:runtime}
\end{wrapfigure}

\paragraph{Runtime Scalability.}
First, we evaluate our algorithm's scalability with respect to team size.
Thanks to its decentralized design, our motion planning policy network's inference time for a single forward pass is \textbf{1.09 ms} (920Hz) with a single CPU thread and constant with respect to the team size.
However, for a fair comparison with an open-loop algorithm like centralized BiRRT, we plot centralized BiRRT's runtime with our approach's total trajectory computation time, which is the sum of computation time for all steps until the targets are reached, with a single CPU thread in Fig.~\ref{fig:runtime}.
The results show that our approach scales efficiently, and is significantly faster than centralized BiRRT exponential runtime (e.g., our policy solves the task 15 times faster in 10 arm tasks).

\begin{wraptable}{R}{.45\textwidth}
  \small
  \vspace{-8mm}
  \centering
  \begin{tabular}{ccccc}                                          \\\toprule
    $N$ & Easy  & Medium & Hard  & Average           \\\midrule
    1   & 0.988 & -      & -     & 0.988             \\
    2   & 0.959 & 0.904  & 0.876 & 0.943             \\
    3   & 0.969 & 0.945  & 0.915 & 0.960             \\
    4   & 0.956 & 0.937  & 0.937 & 0.951             \\\midrule
    \multicolumn{5}{c}{Beyond team size in training} \\\midrule
    5   & 0.948 & 0.935  & 0.914 & 0.943             \\
    6   & 0.947 & 0.915  & 0.900 & 0.938             \\
    7   & 0.924 & 0.901  & 0.894 & 0.918             \\
    8   & 0.909 & 0.907  & 0.920 & 0.909             \\
    9   & 0.912 & 0.928  & 0.864 & 0.910             \\
    10  & 0.908 & 0.904  & 0.881 & 0.905             \\\bottomrule
  \end{tabular}
  \vspace{-2mm}
  \caption{Static Motion Planning Results.}\label{tab:static}
  \vspace{-5mm}
\end{wraptable}

\mypara{Flexibility in Team Size.}
Second, to evaluate the policy's flexibility in team size, we directly test our policy (trained on only 1-4 arm tasks) on 5-10 arm tasks (Table \ref{tab:static}).
On average, our policy achieves $> 90\%$ success for 1-10 arm motion planning tasks.
Due to its decentralization, our motion planner retains roughly the same success rate when more arms are added to the system, as long as the difficulty of the task remains the same.
The decreasing success rate as team size increases is due to our success metric, combined with each arm having an approximately constant probability of success for each difficulty level 
(e.g., if a single-arm success rate is $0.98$, then the success rate for a team size of $N$ on easy tasks is approximately $0.98^N$).


\begin{wraptable}{r}{.45\textwidth}
  \small
  \centering
  \vspace{-7mm}
  \setlength\tabcolsep{3.1pt}
  \begin{tabular}{lcccc}                                                                                             \\\toprule
                    & \multicolumn{4}{c}{Arm count}                                                     \\
                    & 1                             & 2              & 3              & 4               \\\midrule

    No expert       & 0.001                         & 0.000          & 0.000          & 0.000           \\
    No RL (BC)      & 0.403                         & 0.685          & 0.645          & 0.852           \\
    Selfish         & 0.139                         & 0.001          & 0.000          & 0.000           \\
    Individualistic & 0.224                         & 0.023          & 0.000          & 0.000           \\

    \midrule
    Ours            & \textbf{0.988}                & \textbf{0.943} & \textbf{0.960} & \textbf{0.951 } \\\bottomrule
    \vspace{-5mm}
  \end{tabular}
  \caption{Static Motion Planning Ablations. }\label{tab:ablation}
  \vspace{-4mm}
\end{wraptable}

\mypara{Cooperation through Team Rewards. }
Our algorithm achieves multi-arm cooperation with team rewards and coordination by allowing arms to observe other arms.
To validate the cooperative effect of team rewards, we train \textbf{[selfish]} agents with a different reward function, which rewards +0.1 whenever an arm individually reaches its target and penalizes -0.05 whenever it collides, without a special team reward when all arms reach their target poses.
However, if the selfish reward is too small, arms would rather stand still than risk receiving collision penalties, but if it's too large, the dense selfish rewards accumulate to outweigh the collision penalties and lead to high collision rates.
This leads to selfish agents' poor performance in Tab.~\ref{tab:ablation}.

\mypara{Coordination through Observation of Others.}
To validate the coordinative effect of allowing arms to observe other arms, we compare our policy with an \textbf{[individualistic]} baseline, which is trained from scratch without observations of other arms.
Since arms do not observe each other, reach attempts in the presence of other arms likely result in collision penalties, and thus have a lower return than standing still.
Tab. \ref{tab:ablation} shows that our policy outperforms the individualistic baseline, demonstrating that observations of other arms allow for effective coordination.

\mypara{BiRRT Expert Demonstrations for Efficient Exploration.}
With \textbf{[no expert]} demonstrations, the agent rarely experiences positive feedback while frequently collecting penalties due to collisions.
Therefore, the agent learns to avoid collisions by standing still, which results in its 0\% success rates (Tab.~\ref{tab:ablation}).
Additionally, 
a decentralized behavior cloned policy \textbf{[no RL (BC)]} which maximizes the log-likelihood of expert actions given observations from expert observation-action pairs also performs poorly (Tab.~\ref{tab:ablation}), despite having access to the same expert demonstrations.
We hypothesize that this is because of its lack of exposure to failed attempts combined with policy drift, something which our policy sidesteps by actively exploring its environment and learning from both successful and failed experiences.
Interestingly, our policy also completes tasks faster than the behavior cloned policy (Fig. \ref{fig:runtime}), which suggests that our policy improves its optimality through self-exploration beyond the expert experience it receives.
We also observe that the amount of training data supplied to the behavior cloned policy greatly influences its performance.
Since there are approximately 4 times more trajectories from 4 arm tasks than 1 arm tasks, the behavior cloned policy's performance on 4 arm tasks is higher.

\begin{wrapfigure}{r}{0.45\textwidth}
  \vspace{-4mm}
  \includegraphics[trim=23   0   50   35,clip,width=0.45\textwidth]{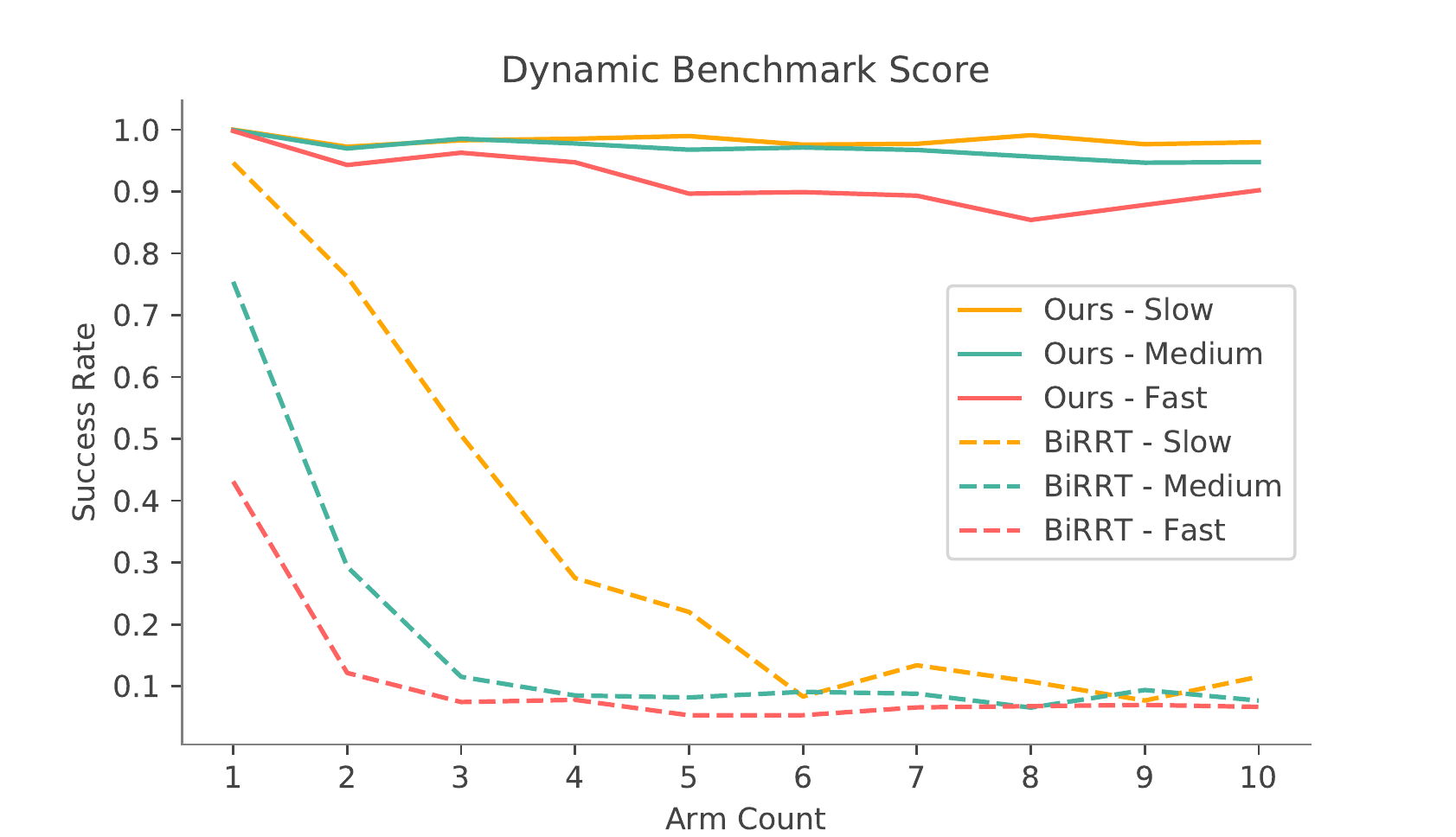}
  \vspace{-6mm}
  \caption{\small The policy directly generalizes to dynamic motion planning tasks, while being trained on static tasks only. Target speed: slow 1-5 cm/s, medium 5-10 cm/s, fast 10-15 cm/s.
  }\label{fig:dynamic}
  \vspace{-5mm}
\end{wrapfigure}

\mypara{Closed-loop Planning for Dynamic Targets.}
In contrast to open-loop motion planners like BiRRT which computes entire trajectories, our closed-loop motion planner only finds the next waypoint which brings it closer to its target in each time step.
Combined with its fast inference speed, our motion planning policy can generalize to dynamic tasks without explicitly modeling its target's dynamics and while being trained only on static tasks.
Fig.~\ref{fig:dynamic} shows the performance in dynamic reaching tasks with three task speed levels.
Our approach's advantage over centralized BiRRT becomes more apparent for faster moving targets and larger team sizes.

\begin{wrapfigure}{r}{0.45\textwidth}
   
  \centering
  \vspace{-5mm}
  \includegraphics[trim=120   0   80   0,clip,width=0.44\textwidth]{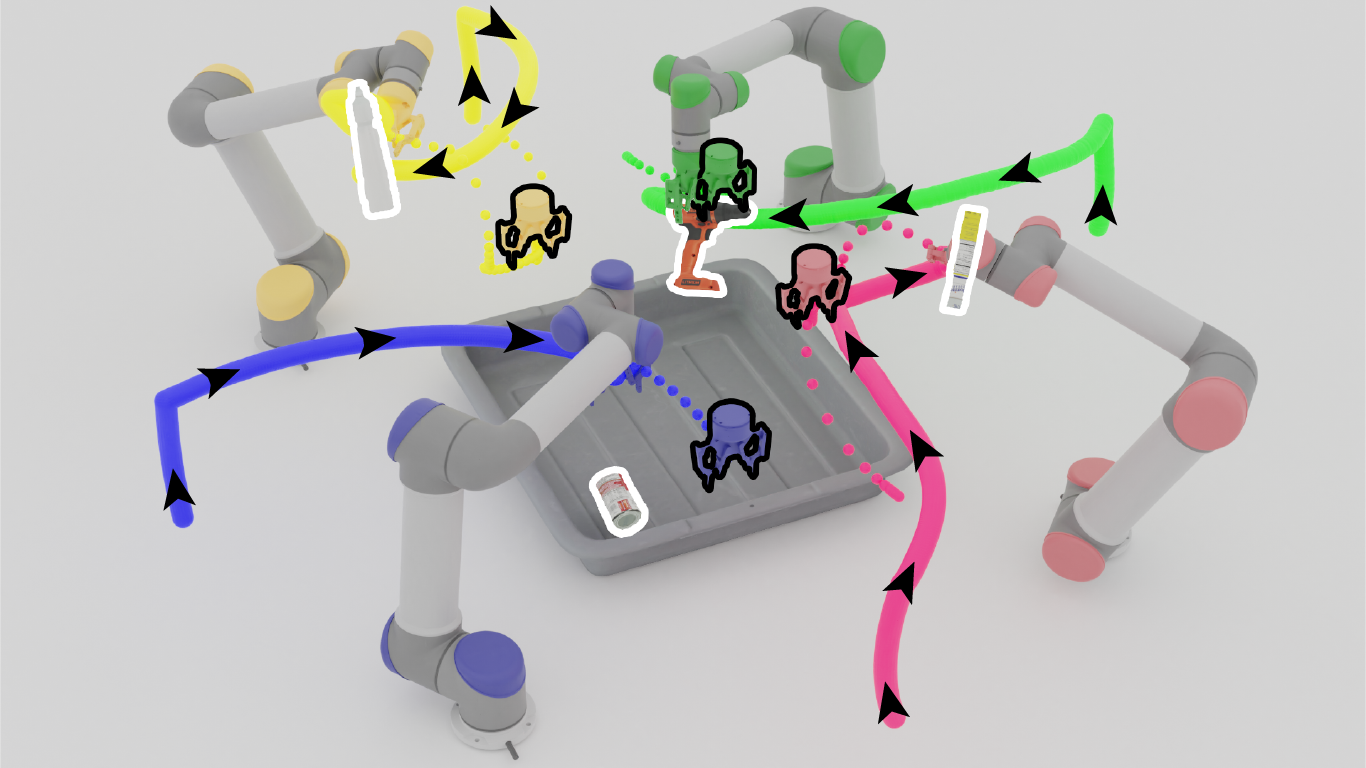} \vspace{-1mm}
  \caption{
    In this multi-arm 6-DoF pick \% place task, arms are controlled to asynchronously pick objects (outlined white) and place them in the bin by supplying target poses (outlined in black) to each arm's policy.
    The arms' past trajectories are solid lines and future trajectories are dotted lines.
  }
  \label{fig:demo}
  \vspace{-6mm}
\end{wrapfigure}

\vspace{-1mm}
\subsection{Application.}\label{section:demo}
\vspace{-2mm}

While multi-arm applications may have arms performing different high-level tasks (e.g: pick \& place, soldering, etc.), all arms rely on the same low-level ability reaching target poses in a collision-free manner. 
When used in conjunction with high-level task planners, our decentralized policy can be extended to any multi-arm team size with the same hardware, to achieve task-level heterogeneity.

We demonstrate our policy's practicality, flexibility, and robustness in a 4-arm 6-DoF pick and place task, where each robotic arm must pick objects from the floor and place them in a bin in the center.
4 robot arms are placed on the corners of a square with side lengths 120cm, with a 72cm $\times$ 77cm $\times$ 10cm bin (Fig \ref{fig:demo}).
We use PyBullet \cite{coumans2017pybullet} and 6-DoF UR5 robots with Robotiq 2F-85 grippers for the demo task.

In each run of our demo environment:
(1) grasp objects are randomly sampled from a YCB dataset~\citep{ycb2015calli} subset consisting of 6 objects plus a cube, and
(2) the initial joint configurations for each arm is uniformly randomized in the range [-0.4rad, 0.4rad] from a home configuration.
Our policy is used to move arms in 3 phases: (1) from their initial joint configurations to the pregrasp pose, (2) from the grasp pose to the dump pose, and (3) from the dump pose its final home configuration.
A low-level controller is used to handle the grasping and dumping of the objects. 
For each object, we precompute and keep 100 most successful grasps from GraspIt!~\citep{miller2004graspit} with noise~\citep{weisz2012pose}.
Pregrasps are computed by applying a 10cm back-off along the approaching direction of grasps.

A demo task is successful if all arms pick and place their target objects in the bin, without colliding with each other or the ground.
We discard runs where the robot knocks over objects or collides with the bin, since arms only observe other arms' states.
To address this limitation, future work could explore leveraging vision systems as perception input to the state encoder.
Averaged over 500 repetitions, our planner achieves \textbf{$78.3\%$} success rate and takes $6272$ steps ($6.8$s) to finish the task.

 \vspace{-3mm} \section{Conclusion} \vspace{-2mm}
We have proposed a method for learning a closed-loop decentralized multi-arm motion planner that scales sub-linearly with respect to the number of arms.
Our experiments demonstrate that the resulting motion planning policy performs well not only in challenging multi-arm motion planning tasks but directly generalizes to tasks with a higher number of arms or tasks with dynamic target end-effector poses, despite being trained only on tasks with static target end-effector poses, while achieving fast closed-loop planning speeds even on a single CPU thread.


\acknowledgments{
We would like to thank Google for the UR5 robot hardware. This work was supported in part by the Amazon Research Award, the Columbia School of Engineering, as well as the National Science Foundation under CMMI-2037101.
}
\setlength{\bibsep}{5pt plus 0.3ex}
\bibliography{references}
\newpage
\begin{center}
    \textbf{\Large{Learning a Decentralized Multi-arm Motion Planner}}
\end{center}

\section{Justification on Sim2Real Transfer}
Our algorithm is tested in the PyBullet simulation environment \cite{coumans2017pybullet}.
We are not able to provide real-world experiments.
However, we believe our algorithm is able to generalize to real-world robot setup for the following reasons:

First, our system uses joint state as input instead of estimation from a perception algorithm. 
In current industry level robot systems, the joint state measurements are often highly accurate and the sim2real difference is negligible.

Second, our benchmark environment takes into account the delay of an inference pass of the motion planning policy. 
This means by the time the motion planner's actions are received and executed on the robots, the observations from which those actions were computed have been outdated by the amount of time which a forward pass takes, which is the case for the real-world robot setup. 
However, since our policy has an inference time of 1.09ms on a single CPU thread, our policy is still able to perform well with this delay.

\section{Training details}

\begin{wraptable}{r}{4.5cm}
\vspace{-3mm}
\begin{tabular}{lc} \toprule
     Hyperparameter& Value \\\midrule
     Actor lr & $0.0005$\\
     Q function lr& $0.001$\\
     Discount Factor $\gamma$ & 0.99 \\
     Exponential Decay $\tau$  & 0.001 \\
     Batch Size & 4096\\
     Warm-up Timesteps & 20,000\\
     Replay Buffer Size & 50,000 \\\bottomrule
\end{tabular}
\caption{Hyperparameters.}
    \label{fig:sac-hyperparams}
\vspace{-6mm}
\end{wraptable}

The state of an arm includes its base pose ($7$), its end-effector pose ($7$), its link positions (($30$ for 10 links), its joint configuration ($6$), and its target end-effector pose ($7$).
One frame of history for all arm state components except for base pose is stacked on top, giving a final arm state vector of size $(7 \times 1) + (7 \times 2) + (30 \times 2) + (6 \times 2) + (7 \times 2) = 107$.

The policy consists of an LSTM state encoder and a MLP motion planner.
The LSTM state encoder has input dimension $107$, hidden dimension $256$, $1$ layer, is single directional, and uses a zero initial hidden state.
The MLP has 3 layers, $[256,128]$,$[128,64]$, and $[64,6]$ where $6$ is the action dimension. 
After each MLP layer is a Hyperbolic Tangent activation function.

The Q function's LSTM shares the same architecture with the policy's LSTM, and its MLP differs only in that its output dimension is 1 and does not have an activation function.

The policy was trained using the curriculum in Tab. \ref{tab:curriculum} on position tolerance $\epsilon_p$ and orientation tolerance $\epsilon_r$, and graduates to the next level when it achieves at least $70\%$ success rate on average in the latest 100 episodes.

The hyperparameters used for Soft Actor Critic are shown in Tab. \ref{fig:sac-hyperparams}.

\section{Behavior Cloned Policy to deal with the Sparse Reward Problem}

\begin{wrapfigure}{r}{0.5\textwidth}
\vspace{-6mm}
    \includegraphics[trim=30   0   50   20,clip,width=0.5\textwidth]{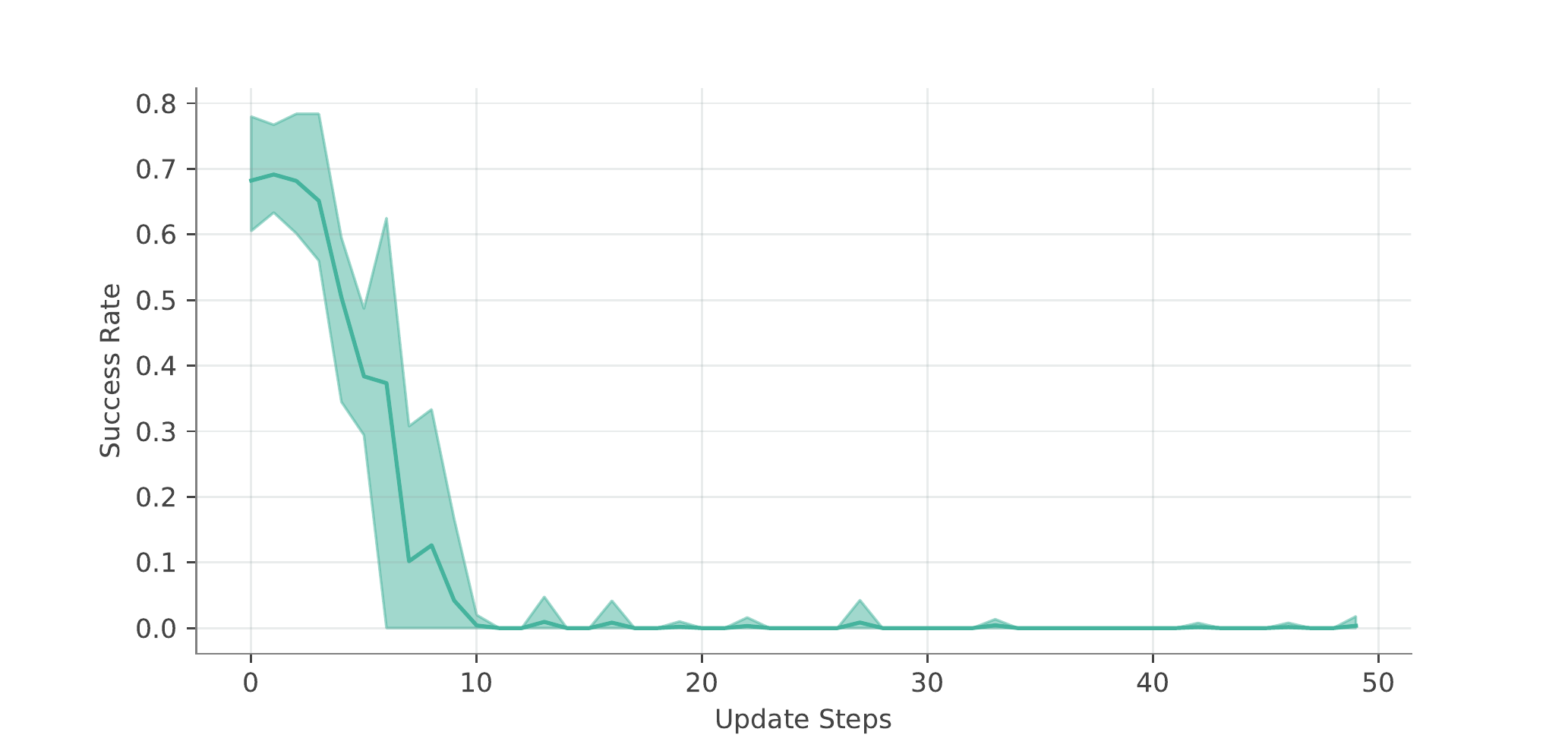}
    \caption{Without expert demonstrations, a behavior cloned policy quickly drops to 0\% success rate. Plot is averaged over 5 seeds.}
    \label{fig:bc-no-ed}
    \vspace{-3mm}
\end{wrapfigure}

While \cite{everett2018ga3ccadrl} could use a pretrained behavior cloned policy for RL in their sparse reward setting, their application was in path planning for grounded robots in a 2D configuration space which corresponds to the cartesian space, which is significantly simpler than motion planning for robotic arms in 6 dimensional joint configuration space.
We observed that a behaviour cloned multi-arm motion planning policy, despite achieving high success rates initially, quickly collapses to 0\% success rate, and is unable to recover, when not provided with expert demonstrations (Fig. \ref{fig:bc-no-ed}).
We hypothesize that for a task as difficult as generic multi-arm motion planning for tightly coupled multi-arm systems, a constant supply of expert demonstrations in the context of failure, is much more helpful for the policy than a good initialization.

\begin{table}
\centering
\begin{tabular}{rrr}
    \toprule
     Level & $\epsilon_p$ (cm) & $\epsilon_r$ (rad) \\\midrule
     1 & 10.0 & 0.20 \\
     2 & 8.0 & 0.16\\
     3 & 6.0&  0.14\\
     4 & 4.0&  0.1\\
     5 & 3.6&  0.09\\
     6 & 3.2&  0.08\\
     7 & 2.8& 0.07 \\
     8 & 2.6&  0.06\\
     9 & 2.4&  0.05\\
     10 & 2.2& 0.05\\
     11 & 2.1& 0.05\\
     12& 2.0& 0.05\\
     13 & 1.9& 0.05\\
     14 & 1.8& 0.05\\
     15 & 1.7& 0.05\\
     16 & 1.6& 0.05\\
     17 & 1.5& 0.05\\
     18 & 1.4& 0.05\\
     19 & 1.3& 0.05\\
     20 & 1.2& 0.05\\
     21 & 1.1& 0.05\\
     21 & 1.0& 0.05\\\bottomrule
\end{tabular}

\caption{Training Curriculum}
\label{tab:curriculum}
\end{table}

\section{Denser Reward Alternative to side-step the Sparse Reward Problem}

\citeauthor{semnani2020multi} \cite{semnani2020multi} addressed \cite{everett2018ga3ccadrl}'s drawback of relying on a pretrained behavior cloned policy with a dense delta-position based reward.
However, their robots are single-linked, while our robot arms have multiple links, which means the arms can easily get stuck in local optima with a corresponding delta end-effector position reward, especially when arms are close to each other.
Thus, such a dense reward scheme would introduce incentives issues, and can not be applied to our problem.
%
%



\end{document}